\documentclass[runningheads]{llncs}

\usepackage{accv}

\usepackage{accvabbrv}

\usepackage{graphicx}
\usepackage{booktabs}
\usepackage{amsmath,amssymb}
\usepackage{tikz}
\usetikzlibrary{arrows.meta,positioning}
\IfFileExists{axessibility.sty}{\usepackage[accsupp]{axessibility}}{}  

\usepackage[pagebackref,breaklinks,colorlinks,citecolor=accvblue]{hyperref}
\IfFileExists{orcidlink.sty}{\usepackage{orcidlink}}{\providecommand{\orcidlink}[1]{}}

\begin{document}

\title{Training-Free Open-Vocabulary 3D Point-Cloud Segmentation\\on the Generalized Few-Shot Benchmark}
\titlerunning{Training-Free Open-Vocabulary GFS-PCS}

\author{Silas Kwabla Gah \and Ebenezer Owusu}
\authorrunning{S.\,K.\,Gah and E.\,Owusu}
\institute{Department of Computer Science, University of Ghana, Legon, Accra, Ghana \\
  \email{skgah001@st.ug.edu.gh} \and \email{ebeowusu@ug.edu.gh}}

\maketitle

\begin{abstract}
Generalized few-shot 3D point-cloud segmentation (GFS-PCS) asks a model to segment a scene into
many base classes seen at training time and a set of novel classes. The state of the art reaches
novel classes by reconciling a dense but noisy 3D vision-language prior with the few-shot
support, but it pays for this with base 3D labels, per-episode training, and the support
annotations themselves. We ask how far the same reconciliation can go with \emph{none} of these:
no training, no 3D labels, and not even the few-shot support. We pair a frozen 3D vision-language
model (RegionPLC) as a dense prior with a frozen promptable concept segmenter (SAM3), prompted by
the bare novel class names and lifted from posed RGB views, and reconcile the two by
\emph{cross-view consistency}: a point becomes novel only when enough of the views that see it
agree. On the ScanNet200 GFS-PCS benchmark this fully training-free, open-vocabulary pipeline
improves novel mIoU by $+2.6$ over the training-free dense prior while holding base
accuracy within $0.5$, and recovers a third ($33\%$) of the novel-class gap to the trained state of
the art that uses far more supervision. We further show that injecting the few-shot support into the pipeline, as a fusion gate and as a
prototypical dense classifier, adds nothing over consistency alone and in fact degrades it through
the classifier, which is why the method needs no support at all. On the harder ScanNet++ benchmark,
where the dense prior is far weaker on novel classes, the same pipeline nearly doubles novel mIoU
($+15.7$, from $16.2$ to $31.9$) at a $1.7$ base cost, lifting the harmonic mean from $21.5$ to
$31.1$.
\end{abstract}

\section{Introduction}
Understanding the semantics of a 3D scene requires naming both the common structures that fill
every room and the long tail of objects that appear rarely and are almost never annotated in 3D.
Generalized few-shot 3D point-cloud segmentation (GFS-PCS) formalizes this regime: a model must
segment a scene into many base classes seen at training time and a set of novel classes each
specified by only a handful of support examples, predicting base and novel jointly at inference
without per-query support \cite{gfsvl}.

The central difficulty of GFS-PCS is that the two sources of novel-class knowledge are
complementary but mismatched. A 3D vision-language model offers a dense prior over the whole
scene, but its predictions are noisy and weak on fine-grained and long-tail classes. The
few-shot support is precise, but it is sparse and covers only a few examples. State-of-the-art
work reconciles the two: GFS-VL \cite{gfsvl} selects high-quality regions from the dense 3D-VLM
pseudo-labels, infills the rest from the support, and mixes novel exemplars into base scenes.
This reconciliation is effective, but it is bought with \emph{training}: GFS-VL pretrains a 3D
segmentor on base 3D ground-truth labels and fine-tunes a registration stage on the support.
Both the base 3D labels and the per-episode optimization are exactly the costs that make
deployment to new label sets and new environments expensive.

\begin{figure}[!ht]
\centering
\includegraphics[width=\columnwidth]{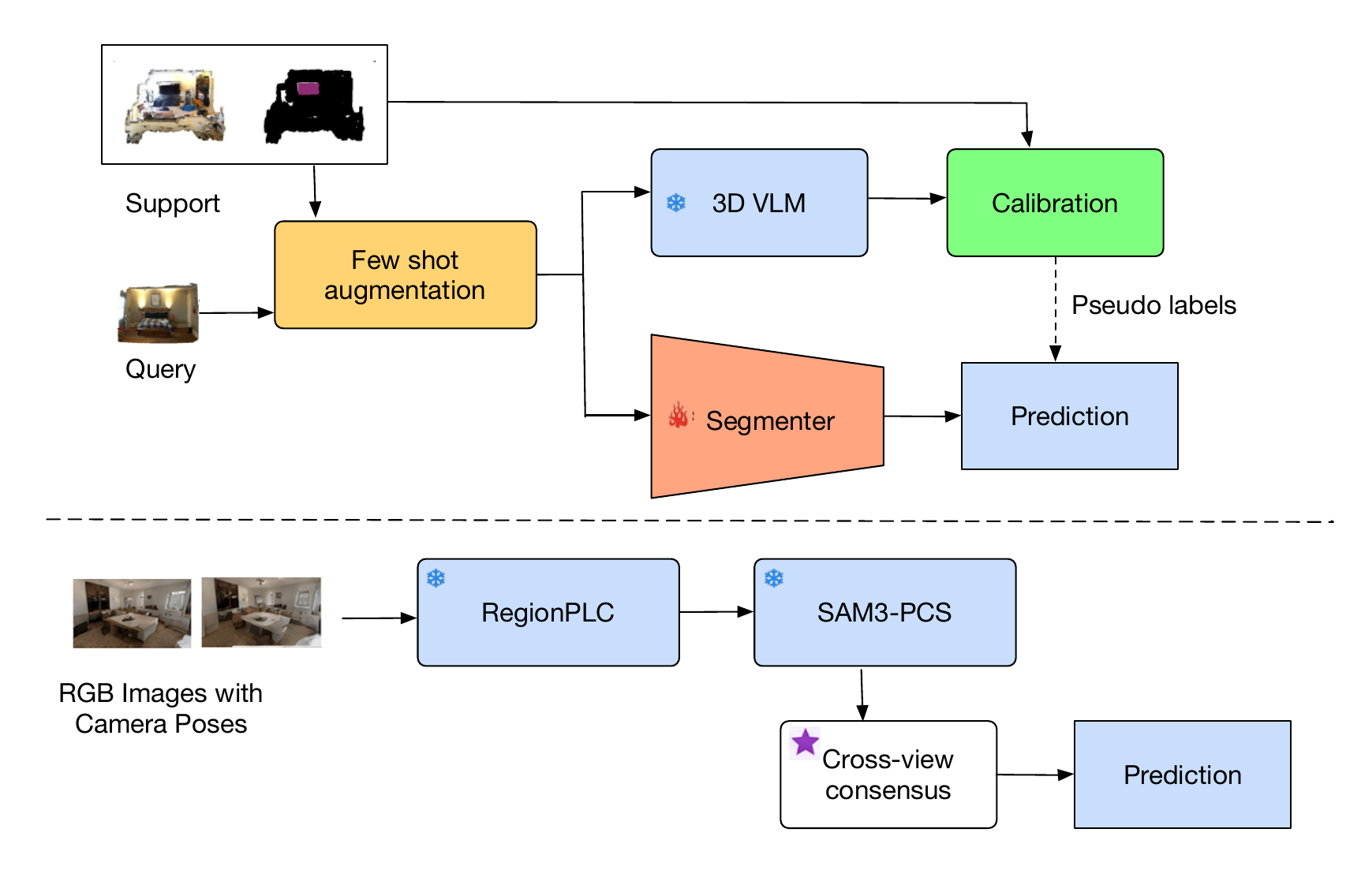}
\caption{\textbf{Prior GFS-PCS vs.\ ours.} \textbf{Top:} the state of the art (GFS-VL) trains a
3D segmentor on base 3D labels and fine-tunes a registration step on the few-shot support.
\textbf{Bottom:} we pair a frozen 3D vision-language model (RegionPLC) with a frozen promptable
concept segmenter (SAM3-PCS) prompted by the bare novel class names, and reconcile their
predictions by \emph{cross-view consistency}. We use no training, no 3D labels, and not even the
few-shot support, yet recover $+15.7$ novel mIoU over the training-free dense prior.}
\label{fig:teaser}
\end{figure}

We ask a different question: how far can the dense-sparse reconciliation go with \emph{no
training and no 3D labels of any kind}? We answer it with two frozen foundation models and a
parameter-free rule (Fig.~\ref{fig:teaser}). A frozen 3D vision-language model (RegionPLC \cite{regionplc}) provides the
dense prior. A frozen promptable concept segmenter (SAM3 \cite{carion2025sam3segmentconcepts}), prompted by the bare
novel class names, provides the sparse source: we run it on the scene's posed RGB frames and
lift its masks to the point cloud by projection. Nothing is trained, no 3D annotation is used,
and the few-shot support is never consumed; the only inputs are the class names and the RGB-D
capture that already produced the point cloud.

The naive way to combine them, letting any lifted novel mask overwrite the dense prediction,
fails: the 2D segmenter over-fires onto base surfaces, so novel accuracy rises only by
sacrificing base accuracy. Our key finding is that \emph{cross-view consistency} resolves this.
A point is relabeled novel only when a sufficient fraction of the views that see it agree on
that novel class. Consistency suppresses single-view false positives, which raises novel
accuracy while leaving base accuracy intact. Consistency turns out to be the whole story: three
natural alternatives that gate the lifted votes by view-count visibility, by dense-model
confidence, or by similarity to feature prototypes built from the few-shot support each add
nothing over consistency alone. The last of these is the reason we report an open-vocabulary
method: once consistency is applied, the few-shot support carries no additional signal, so the
reconciliation needs a single threshold, no learned parameters, and no support annotations.

On the ScanNet200 GFS-PCS benchmark our fully training-free pipeline raises novel mIoU by
$+2.6$ over the training-free dense prior at a negligible $0.5$ base cost, recovering a third
($33\%$) of the novel-class gap to trained GFS-VL. On the harder ScanNet++ benchmark, whose dense
prior is much weaker on novel classes, the same pipeline raises novel mIoU by $+15.7$ (from $16.2$
to $31.9$) at a $1.7$ base cost, lifting the harmonic mean from $21.5$ to $31.1$.

\paragraph{Contributions.}
\begin{itemize}
  \item A \emph{fully training-free, open-vocabulary} method for the GFS-PCS benchmark: frozen
        RegionPLC (dense) and frozen SAM3-PCS (sparse, lifted from posed RGB) reconciled with no
        learned parameters, no base 3D labels, no fine-tuning, and no few-shot support. It uses
        strictly less supervision than prior GFS-PCS work.
  \item Cross-view consistency as the reconciliation rule: it lifts novel-class accuracy while
        preserving base accuracy, where naive sparse-over-dense overwrite trades one for the other.
        We show that view-count-visibility and dense-confidence gates add nothing over consistency
        alone, and, more strongly, that injecting the $K$ few-shot support samples at the two loci
        our frozen text-conditioned pipeline admits (fusion arbitration and dense classification)
        never improves novel accuracy: the gate is null within a pre-committed confidence interval
        and the prototypical classifier is significantly negative. Conditioned on consensus, the
        provided support carries no beneficial signal, which is what licenses the support-free,
        open-vocabulary setting.
  \item On ScanNet200 and ScanNet++ we establish the training-free, support-free floor for
        the GFS-PCS benchmark and characterize the remaining gap to the heavily supervised state
        of the art.
\end{itemize}

\section{Related Work}
\paragraph{Few-shot and generalized few-shot 3D segmentation.}
Few-shot 3D point-cloud segmentation (FS-PCS) learns novel classes from a handful of support
samples and predicts only those classes at inference \cite{attmpti,coseg,mmfspcs}. The generalized
setting (GFS-PCS) instead predicts base and novel jointly without per-query support, following its
2D counterparts \cite{capl,pifs}. For 3D, GW \cite{gw} encodes shared geometric structures as
``geometry words'' to enhance prototypes, query-guided enhancement \cite{queryguided} and
pseudo-embedding \cite{pseudoembed} refine the support representation, and non-parametric networks
avoid per-episode training altogether \cite{notimetotrain}; prototype learning \cite{protonet}
underlies most of them. The state of the art, GFS-VL \cite{gfsvl}, reconciles dense, noisy 3D
vision-language pseudo-labels with the sparse, precise few-shot support through pseudo-label
selection, adaptive infilling, and a base-novel mix, and introduces the ScanNet200
\cite{scannet200} and ScanNet++ \cite{scannetpp} GFS-PCS benchmarks. It pretrains a 3D segmentor
(PTv3 \cite{ptv3}) on base 3D ground truth and fine-tunes a registration stage on the support. We
adopt its benchmark and protocol but remove all training and all 3D supervision, and find the
support itself dispensable.

\paragraph{Open-vocabulary 3D scene understanding.}
A line of work distills 2D vision-language features \cite{clip} into 3D for annotation-free
open-vocabulary prediction: OpenScene \cite{openscene} aligns 3D points to CLIP pixel features,
PLA \cite{pla} and Lowis3D \cite{lowis3d} build point-language pairs from image captions, RegionPLC
\cite{regionplc} adds regional point-language contrastive learning, and others lift dense CLIP
features \cite{clipfo3d,clip2scene,clip2point} or align across modalities
\cite{li2024dense}. Open-vocabulary instance segmentation \cite{openmask3d} and
segmentation with 2D foundation models \cite{ovfm3d} extend the idea. These dense priors are broad
but noisy on fine-grained and long-tail classes. We use RegionPLC unchanged as our dense prior; our
contribution is the training-free sparse branch and the reconciliation, not the dense model.

\paragraph{Training-free 2D-to-3D lifting and multi-view consistency.}
The idea of lifting 2D foundation-model outputs to 3D and resolving them across views is shared
by a recent line of \emph{training-free} open-vocabulary pipelines, but that line is
overwhelmingly \emph{instance}-centric. OpenIns3D \cite{openins3d} renders synthetic views,
detects with a 2D open-vocabulary model, and looks category names back up onto class-agnostic 3D
mask proposals; OV-SAM3D \cite{ovsam3d} grows superpoints into 3D prompts refined by SAM masks
for training-free instance understanding. These methods confirm that cross-view aggregation of 2D
priors is powerful without 3D training, but they target class-agnostic mask proposals plus a
naming lookup, not per-point \emph{semantic} labelling, and they do not reconcile a dense 3D
vision-language prior with the lifted evidence. Adapting an instance-proposal pipeline to report
per-point semantic mIoU on the fixed base/novel split of the GFS-PCS protocol is non-trivial and
would change the comparison's supervision profile; we therefore take RegionPLC zero-shot, the
dense prior our method builds on, as the directly comparable training-free baseline, and treat
the instance-centric methods as orthogonal evidence for the value of multi-view consistency rather
than as drop-in semantic baselines. A complementary family trains dense 3D encoders with large-scale multimodal supervision, DMA \cite{li2024dense} and UniM-OV3D \cite{unimov3d} co-embed
point, image, text (and depth) features, Mosaic3D \cite{mosaic3d} pretrains a 3D foundation model
on millions of region-caption pairs, and SOLE \cite{sole} learns language-conditioned 3D instance
masks and reaches substantially higher open-vocabulary accuracy, but at exactly the training and
data cost our setting forgoes. Any of these stronger dense models could in principle replace
RegionPLC as our prior; doing so would raise the floor we report without changing the reconciliation,
and we leave that substitution to future work.

\paragraph{2D vision-language and promptable foundation models.}
Image-text pre-training \cite{clip,flamingo,kosmos2} enables open-vocabulary 2D recognition, and a
family of grounded dense-prediction models extends it to localization: language-driven segmentation
\cite{lseg}, open-vocabulary segmentation from image-level labels \cite{openseg}, dense labels from
frozen CLIP \cite{maskclip}, and open-world detection \cite{li2022grounded,detic,detclip}. SAM \cite{sam} and
SAM2 \cite{sam2} segment any region from geometric prompts but are class-agnostic; SAM3 \cite{carion2025sam3segmentconcepts}
adds promptable concept segmentation, returning all instances of a named or exemplared concept. We
use SAM3 concept segmentation as a precise, training-free source of novel-class evidence in 2D and
lift it to 3D by projection.

\paragraph{3D backbones, benchmarks, and augmentation.}
Our dense prior builds on sparse-convolutional and transformer point-cloud backbones
\cite{sparseconv,minkowski,kpconv,pointconv,stratified,ptv3}, and we evaluate on standard indoor
benchmarks \cite{scannet,scannet200,scannetpp,s3dis} (GFS-VL also reports outdoor nuScenes
\cite{nuscenes}). Unlike GFS-VL, which mixes novel support objects into base scenes in the spirit of
3D data augmentation \cite{mix3d,pointmixup}, we use the support only at inference, if at all.

\section{Problem Setup}
We adopt the GFS-PCS protocol and benchmarks of GFS-VL \cite{gfsvl}. Let $\mathcal{C}^{b}$ be
the base class set of size $N_b$ and $\mathcal{C}^{n}$ the novel class set of size $N_n$, with
$\mathcal{C}^{b}\cap\mathcal{C}^{n}=\varnothing$. For each novel class $c$ the supervision is $K$
support samples $\{(\mathbf{X}^{c}_{k},\mathbf{Y}^{c}_{k})\}_{k=1}^{K}$. At inference the model labels every point of a test scene over $\mathcal{C}^{b}\cup\mathcal{C}^{n}$ without per-query
support, and is scored by mean IoU over base, over novel, and their harmonic mean. On ScanNet200
$N_b{=}12,\,N_n{=}45$ and on ScanNet++ $N_b{=}12,\,N_n{=}18$.

Unlike GFS-VL we never use the base 3D labels, never train, and never consume the $K$ support
samples. Our only supervision is the novel \emph{class names}, which name the concepts the
promptable sparse segmenter looks for (Sec.~\ref{sec:method}); the dense prior is likewise
zero-shot. This places our method in the GFS-PCS evaluation protocol while using strictly less
supervision than it allows. Section~\ref{sec:abl} shows that adding the few-shot support back, as
feature prototypes that gate the lifted votes, does not change the result, which is what licenses
the open-vocabulary setting.
\begin{figure}[!ht]
\centering
\includegraphics[width=\columnwidth]{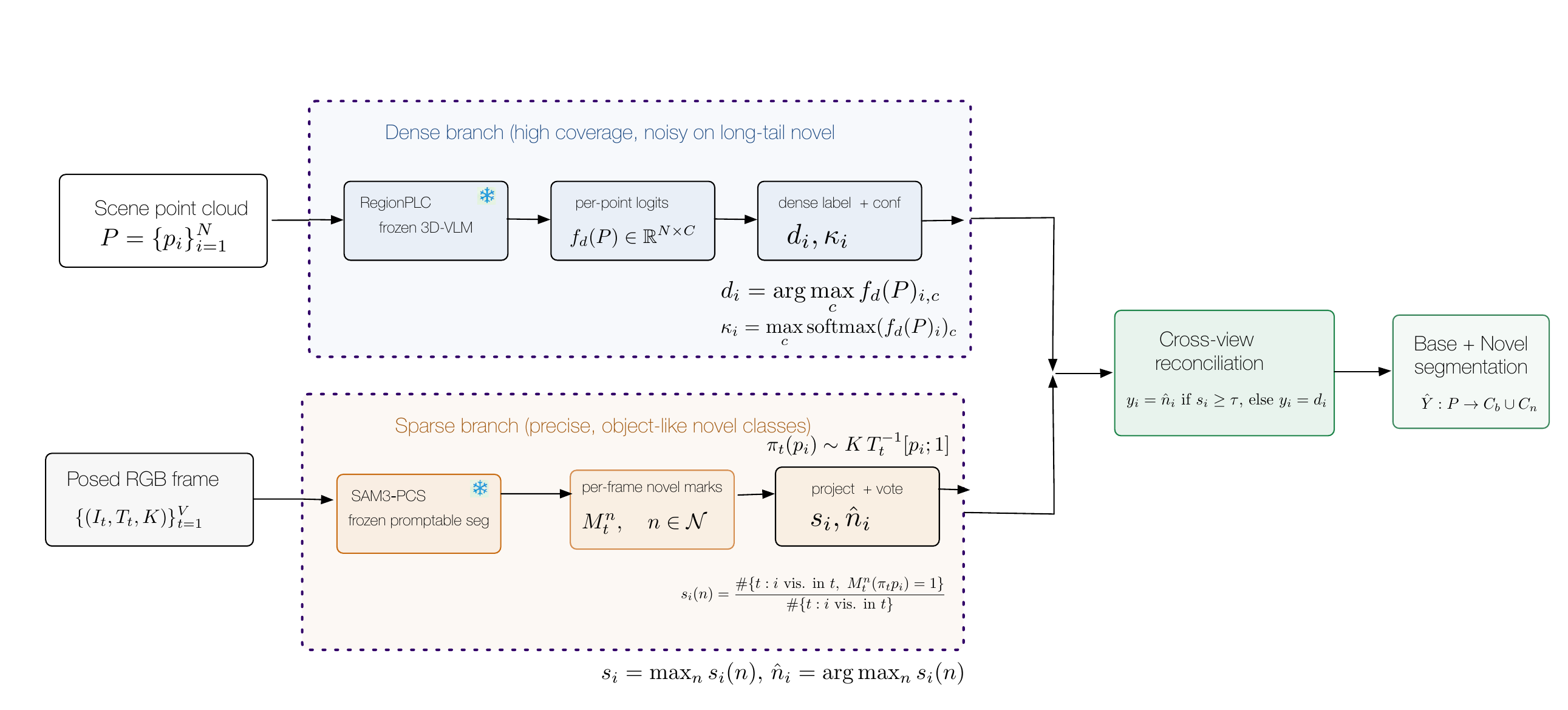}
\caption{\textbf{Pipeline overview (all components frozen).} A frozen 3D-VLM gives a dense but
noisy per-point prediction $d_i$. SAM3-PCS, run on the posed RGB frames with the bare novel class
names and lifted to the point cloud, gives a \emph{cross-view agreement} $s_i$ for the winning
novel class. Reconciliation overwrites the dense label with the lifted novel label only where the
agreement clears the single threshold $\tau$ (Sec.~\ref{sec:method}). Nothing is trained; the
few-shot support is never used.}
\label{fig:arch}
\end{figure}
\section{Method}\label{sec:method}

Our pipeline has three training-free stages (Fig.~\ref{fig:arch}) over a scene point cloud
$P=\{p_i\}_{i=1}^{N}$ and its posed RGB frames $\{(I_t, T_t, K)\}$  ($T_t$ the camera-to-world
pose, $K$ the intrinsics).

\paragraph{Dense branch.}
A frozen 3D vision-language model $f_d$ (RegionPLC) maps the point cloud to per-point
open-vocabulary logits over the 57 evaluated classes. We take the per-point dense label
$d_i=\arg\max_c f_d(P)_{i,c}$ and confidence $\kappa_i=\max_c \mathrm{softmax}(f_d(P)_i)_c$.
This branch has high coverage but is noisy on long-tail novel classes.

\paragraph{Sparse branch (SAM3-PCS lifted).}
For each novel concept $n\in\mathcal{N}$ and frame $I_t$ we run SAM3 promptable concept
segmentation to obtain a 2D mask $M_t^n$. We project every point into each frame,
$[u,v,z]^\top \sim K\, T_t^{-1}[p_i;1]$, keep points with $z>0$ inside the image, and read the
predicted novel concept at the pixel. Aggregating over all frames a point is visible in, we form
the \emph{cross-view agreement}
\begin{align*}
  s_i(n) &= \frac{\#\{t: \text{point }i\text{ visible in }t \text{ and } M_t^{n}(u,v)=1\}}
                 {\#\{t: \text{point }i\text{ visible in }t\}}, \\[4pt]
  s_i &= \max_n s_i(n), \qquad \hat n_i = \arg\max_n s_i(n).
\end{align*}
$s_i$ is itself a training-free confidence: a point whose 2D novel label is consistent across
many views is reliable; a single-view false positive is suppressed.

\paragraph{Reconciliation.}
We overwrite the dense label with the sparse novel label only where cross-view agreement clears
a threshold $\tau$:
\[
  y_i = \begin{cases} \hat n_i & s_i \ge \tau\\ d_i & \text{otherwise.}\end{cases}
\]
The consistency threshold $\tau$ is the single hyper-parameter. We \emph{pre-commit} it per
benchmark to a conservative round value rather than the eval-optimal one---$\tau{=}0.5$ on
ScanNet200 and $\tau{=}0.15$ on ScanNet++---so the operating point is never tuned to maximise the
reported metric (the selection protocol and a leakage discussion are given in the supplementary).
Because $s_i$ is a fraction of agreeing
views, its numeric scale depends on how densely each capture samples viewpoints, so the operating
point is calibrated per benchmark rather than transferred as a raw constant
(Sec.~\ref{sec:abl}). Section~\ref{sec:abl} further shows that gating additionally on dense
uncertainty $\kappa_i$ does not improve over consistency alone, so the reconciliation is
parameter-free beyond $\tau$.

\section{Experiments}
\paragraph{Datasets and metrics.} ScanNet200 GFS-PCS (12 base / 45 novel; 312 validation scenes)
and ScanNet++ GFS-PCS (12 base / 18 novel; 50 validation scenes, 48 of which release posed DSLR
imagery used by the sparse branch). Mean IoU over base, novel, all classes, and the base-novel
harmonic mean, all training-free and label-free.
\paragraph{Evaluation protocol.} We report on the full ScanNet200 validation set (312 scenes).
The dense baseline is evaluated over all scenes. The sparse branch additionally requires the
posed RGB frames of each scene, which we obtain from the ScanNet sensor stream; the lifted
predictions are evaluated on the same full validation set, so the dense, fused, and trained-SOTA
numbers are directly comparable. The headline fused number is over all 312 scenes; the threshold
$\tau$ is fixed once per benchmark and never tuned on the evaluation scenes (selection protocol in
the supplementary).
\paragraph{Baselines.} Dense-only (RegionPLC zero-shot, the training-free floor); naive
sparse-over-dense (any lifted novel mask overwrites the dense label, operationally the lowest
threshold in our sweep, $\tau{=}0.2$, with no further consistency filtering); and trained GFS-VL
(upper reference, NOT a beat-target).
\paragraph{Implementation.}
\emph{Dense prior.} We use RegionPLC \cite{regionplc} with a SparseUNet32 backbone and the
annotation-free (zero-shot) weights released by GFS-VL \cite{gfsvl}, unchanged and frozen; it
produces per-point open-vocabulary logits over the evaluated classes.
\emph{Sparse branch.} The promptable segmenter is SAM3 \cite{carion2025sam3segmentconcepts} concept segmentation (frozen),
prompted \emph{one novel class at a time with the benchmark's class name used verbatim}: we apply
no synonyms, no composite or templated prompts, and no prompt engineering, so the only label-side
input is the official class string (we return to this label-list dependence in the discussion). We
keep, per frame and per concept, the highest-scoring returned mask at SAM3 confidence threshold
$0.3$; where masks of different concepts claim the same pixel the higher-scoring concept wins, and
the surviving novel class id is written to a per-frame label image.
\emph{Posed frames.} ScanNet200 frames come from the ScanNet sensor stream (\texttt{.sens}) used to build the point cloud; ScanNet++ has no such stream, so we use its released resized
\emph{undistorted} DSLR images (a pinhole model, no distortion to undo) with intrinsics and
camera-to-world poses from the nerfstudio transforms, converted from OpenGL to OpenCV camera axes.
Both expose poses in the same mesh frame as the point cloud, so a point projects by the plain
pinhole equation. We subsample frames uniformly (every $20^{\text{th}}$ frame, $\approx\!120$ per
scene) and accumulate cross-view agreement over every frame a point falls inside ($V_{\min}{=}1$).
\emph{Threshold.} The single hyper-parameter $\tau$ is fixed once per benchmark and never tuned to
maximise the evaluation metric. Because $s_i$ is a \emph{fraction} of agreeing views, its scale is
set by how densely each capture samples viewpoints---ScanNet200's dense video stream yields high
agreement ($\tau{=}0.5$), ScanNet++'s sparser wide-baseline DSLR photography much lower
($\tau{=}0.15$)---so we calibrate one threshold per benchmark rather than transfer a constant. On
both benchmarks we pre-commit a conservative value \emph{below} the sweep optimum, within a broad
plateau where the harmonic mean varies by under $1.5$ points across the plausible range; the full
selection protocol---including the absence of a separate ScanNet++ development split and why the
single in-set probe scene cannot inflate the result---is given in the supplementary material.
\emph{Compute.} All experiments run on a single NVIDIA T4; nothing is trained or fine-tuned.

The SAM3 weights are publicly available at
\url{https://huggingface.co/facebook/sam3} in a gated
access model that requires a one-time request approval.

\begin{table}[t]\centering
\caption{ScanNet200 GFS-PCS on the full validation set (312 scenes). We follow the metric
convention of GFS-VL \cite{gfsvl}: mean IoU over base (mIoU-B), novel (mIoU-N), all classes
(mIoU-A), and the base-novel harmonic mean (HM). Our method uses \emph{no} support, so it has a
single setting; trained GFS-VL is shown at its strongest (5-shot) as an upper reference, not a
beat-target.}
\begin{tabular}{lcccc}
\toprule
Method & mIoU-B & mIoU-N & mIoU-A & HM \\
\midrule
Dense only (RegionPLC, zero-shot)     & 46.93 & 23.78 & 28.65 & 31.56 \\ 
Naive sparse-over-dense               & 44.39 & 24.51 & 28.70 & 31.58 \\ 
\textbf{Ours} (cross-view consistency) & 46.44 & 26.42 & 30.63 & \textbf{33.68} \\ 
\midrule
GFS-VL \cite{gfsvl} (trained, 5-shot, upper ref.) & 67.57 & 31.67 & 39.23 & 43.12 \\
\bottomrule
\end{tabular}
\end{table}

\begin{table}[t]\centering
\caption{ScanNet++ GFS-PCS on all $50$ validation scenes; the two scenes without posed DSLR imagery
fall back to the dense prediction in both training-free rows, so dense and fused remain directly
comparable over the same scene set. $12$ base /
$18$ novel, the same metric convention and upper reference as the ScanNet200 results above. GFS-VL
does not report a training-free RegionPLC zero-shot row on ScanNet++, so this dense floor is a new
reference point. The operating point $\tau{=}0.15$ is fixed on a single probe scene (ScanNet++
provides no separate development split; see supp.) and is conservative---below the eval optimum---and
lower than ScanNet200's $0.5$ because ScanNet++'s sparse, wide-baseline DSLR photography yields
lower cross-view agreement than ScanNet200's dense video stream (Sec.~\ref{sec:method}).}
\label{tab:scannetpp}
\begin{tabular}{lcccc}
\toprule
Method & mIoU-B & mIoU-N & mIoU-A & HM \\
\midrule
Dense only (RegionPLC, zero-shot)     & 32.10 & 16.16 & 22.54 & 21.50 \\ 
\textbf{Ours} (cross-view consistency) & 30.37 & 31.91 & 31.29 & \textbf{31.12} \\ 
\midrule
GFS-VL \cite{gfsvl} (trained, 5-shot, upper ref.) & 60.05 & 21.66 & 37.02 & 31.82 \\
\bottomrule
\end{tabular}
\end{table}

\paragraph{Statistical significance.} The novel-mIoU gains are not artifacts of a few scenes. A
paired bootstrap over the evaluation scenes (per-scene novel mIoU, $10^4$ resamples) gives, on
ScanNet200, $\Delta\text{mIoU-N}{=}{+}2.6$ with a $95\%$ CI of $[+2.13,+3.23]$, and on ScanNet++
$\Delta\text{mIoU-N}{=}{+}15.7$ with a $95\%$ CI of $[+13.17,+19.42]$; both intervals exclude zero.

\section{Ablations}\label{sec:abl}

\begin{table*}[t]\centering
\caption{\textbf{Ablations on ScanNet200 GFS-PCS} (training-free). mIoU-B/N = base/novel mean IoU;
HM = base-novel harmonic mean; $\tau$ = cross-view consistency threshold (pre-committed
$\tau{=}0.5$). Blocks (a--c) are full validation (312 scenes); block (d) is the 26-scene
development set. $\Delta$mIoU-N in (d) is the change in novel mIoU vs.\ the support-free reference
with a $95\%$ paired-bootstrap CI; a CI bracketing $0$ means the locus is null.}
\label{tab:ablation}

\begin{minipage}[t]{0.46\linewidth}\centering
\textbf{(a) Reconciliation rule} ($\tau{=}0.5$, full val)\\[3pt]
\footnotesize
\begin{tabular}{lccc}
\toprule
 & mIoU-B & mIoU-N & HM \\
\midrule
Dense only (RegionPLC)       & 46.93 & 23.78 & 31.56 \\
Naive sparse-over-dense      & 44.39 & 24.51 & 31.58 \\
\textbf{Ours} (consistency)  & 46.44 & 26.42 & \textbf{33.68} \\
\bottomrule
\end{tabular}
\end{minipage}\hfill
\begin{minipage}[t]{0.46\linewidth}\centering
\textbf{(b) Consistency threshold $\tau$} (full val)\\[3pt]
\footnotesize
\begin{tabular}{cccc}
\toprule
$\tau$ & mIoU-B & mIoU-N & HM \\
\midrule
0.2 & 44.39 & 24.51 & 31.58 \\
0.3 & 45.67 & 25.86 & 33.02 \\
0.4 & 46.19 & 26.35 & 33.56 \\
0.5 & 46.44 & 26.42 & 33.68 \\
0.6 & 46.75 & 26.39 & 33.74 \\
0.7 & 46.88 & 26.06 & 33.49 \\
\bottomrule
\end{tabular}
\end{minipage}
\medskip

\begin{minipage}[t]{0.46\linewidth}\centering
\textbf{(c) Auxiliary gates over consistency} ($\tau{=}0.5$)\\[3pt]
\resizebox{\linewidth}{!}{%
\begin{tabular}{lccc}
\toprule
Gate & mIoU-B & mIoU-N & HM \\
\midrule
none (consistency only)   & 46.44 & 26.42 & \textbf{33.68} \\
$+$ view-count visibility  & 46.44 & 25.94 & 33.29 \\ 
$+$ dense-confidence       & 46.54 & 26.18 & 33.51 \\ 
\bottomrule
\end{tabular}}
\end{minipage}\hfill
\begin{minipage}[t]{0.46\linewidth}\centering
\textbf{(d) Support injection: is support necessary?}\\[3pt]
\resizebox{\linewidth}{!}{%
\begin{tabular}{lcc}
\toprule
Locus & $\Delta$mIoU-N & 95\% CI \\
\midrule
support-free (reference)    & --      & -- \\
L1 fusion gate              & $+0.03$ & $[+0.00,+0.12]$ \\
L2 prototypical logits      & $-4.89$ & $[-7.92,-1.45]$ \\
\bottomrule
\end{tabular}}\\[3pt]
{\footnotesize $26$ dev scenes, $\tau{=}0.5$, $10^4$-resample paired bootstrap. L1 at its best
threshold (sim$\,\ge\!0.3$) brackets zero (null); L2 at its least-harmful blend ($\alpha{=}0.25$)
excludes zero but is \emph{negative} and worsens with $\alpha$. L3 (exemplar) not instantiated;
see Sec.~\ref{sec:support}.}
\end{minipage}
\end{table*}

\paragraph{Cross-view consistency is necessary.} Without it (naive sparse-over-dense overwrite,
at $\tau{=}0.2$, the lowest threshold tested) the 2D segmenter's over-fire onto base surfaces drops base mIoU sharply; raising
$\tau$ recovers base and improves novel together: base climbs monotonically from $44.39$ at
$\tau{=}0.2$ to $46.88$ at $\tau{=}0.7$, novel peaks at $\tau{=}0.5$ ($26.42$), and the harmonic
mean is flat to within $0.1$ across $\tau\in[0.5,0.6]$ (Table~\ref{tab:ablation}b). We pre-commit
to $\tau{=}0.5$; the empirical HM maximum at $\tau{=}0.6$ ($33.74$ vs.\ $33.68$) is within noise
and we do not tune $\tau$ on the evaluation scenes.
\paragraph{Visibility and dense-confidence gates add nothing.} Gating the lifted votes additionally
by view-count visibility (a point is relabeled only if seen in at least $V{=}2$ views) leaves base
\emph{identical} and lowers novel by $0.48$; gating by dense-model softmax confidence $\kappa_i$ (a
point is relabeled only when the dense model is unconfident) moves base by $+0.10$ and novel by
$-0.24$. Each leaves the harmonic mean \emph{below} consistency alone ($33.29$ and $33.51$ vs.\
$33.68$, Table~\ref{tab:ablation}c). Cross-view agreement $s_i$
already encodes both signals: a point that many views agree on is both well seen and confidently
named, so an explicit visibility or confidence gate is redundant. The reconciliation is therefore
irreducible to a single threshold $\tau$.

\subsection{Is the few-shot support necessary?}\label{sec:support}
The GFS-PCS protocol provides $K$ labelled support samples per novel class, and prior work treats
them as the primary source of novel-class evidence \cite{gfsvl,protonet}. Our method never
consumes them. To test whether they would help, we do not try a single mechanism and report a
single number; we observe that a frozen dense-plus-sparse pipeline exposes three loci at which
support can enter, and we inject it at the two that our text-conditioned pipeline admits, using the
canonical few-shot operator for each. Let $\phi_i$ denote the $\ell_2$-normalized RegionPLC feature
at point $i$, and let $\mathbf{q}_n$ be the support prototype of novel class $n$: the
$\ell_2$-normalized mean of the frozen RegionPLC features over the $K$ support point sets of class
$n$. The injections are:

\textbf{(L1) Fusion arbitration.} A lifted vote for $n$ at point $i$ is admitted only when
$\langle \phi_i,\mathbf{q}_n\rangle \ge \theta$. This is the metric-matching operator applied
where the sparse branch overwrites the dense one; it filters votes but cannot create them.

\textbf{(L2) Dense classification.} We blend the support prototype into the dense novel score
\emph{before} reconciliation,
\[
  \tilde f_d(P)_{i,n} \;=\; (1-\alpha)\,f_d(P)_{i,n} \;+\; \alpha\,\langle \phi_i, \mathbf{q}_n\rangle,
  \qquad \alpha\in[0,1],
\]
recovering the open-vocabulary dense branch at $\alpha{=}0$ and a prototypical few-shot classifier
\cite{protonet} at $\alpha{=}1$. Unlike L1, large $\alpha$ can manufacture novel predictions the
text prior missed.

\textbf{(L3) Sparse generation (not instantiated).} The third locus would condition the 2D
segmenter on visual exemplars of the support rather than on the class name, and is the one locus
that routes support through a different model and modality. It requires an exemplar-capable
promptable segmenter; the SAM3 concept-segmentation interface we use accepts text and geometric
prompts only, with no cross-image exemplar path, so we cannot instantiate L3 without changing the
sparse model. We leave exemplar-conditioned generation to future work and, because both tested
loci route the support through RegionPLC features, treat the residual possibility of a
shared-feature explanation as a stated limitation.

We evaluate each tested injection against its support-free counterpart on a held-out development
set, and report the mean per-scene change in novel mIoU with a $95\%$ paired-bootstrap confidence
interval ($10^4$ resamples) at the operating $\tau$. The decision rule is pre-committed: an
injection is declared null if its interval contains zero.

Neither operator improves over cross-view consistency, and they fail in two different ways.
L1 metric gating is null: its largest effect over the threshold sweep is
$\Delta\text{mIoU-N}{=}{+}0.03$ (CI $[+0.00,+0.12]$), an interval that brackets zero, because the
gate can only filter lifted votes and the consensus filter has already removed the unreliable ones.
L2 prototypical logit fusion does \emph{not} bracket zero, but its effect is strictly negative
across the entire blend: the least-harmful setting ($\alpha{=}0.25$) already costs
$\Delta\text{mIoU-N}{=}{-}4.89$ (CI $[-7.92,-1.45]$) and larger $\alpha$ is worse, collapsing the
dense novel score from $18.79$ at $\alpha{=}0$ to $0.00$ at $\alpha{\ge}0.5$ as the weakly
calibrated prototype similarity overwhelms the text logits in the model's own score space. The
pre-committed positive-gain outcome, which would have forced us to revisit the support-free framing,
does not occur at either locus: support is statistically indistinguishable from zero through the
gate and actively harmful through the classifier. We therefore state the scoped conclusion the
experiment licenses: in the training-free, open-vocabulary regime, once the dense prior is
reconciled by cross-view consensus, the provided support set yields no positive gain through either
injection mechanism we test. We do not claim that support is universally
useless; trained registration still benefits from it \cite{gfsvl}, and our base-class gap to that
method is the standing cost of forgoing it. Because both tested loci consume support through
RegionPLC features, we cannot fully exclude a shared-feature explanation; the exemplar locus (L3)
that would settle it is left to future work. The claim is narrower and harder to refute:
conditioned on consensus, the marginal value of the \emph{provided} support, injected through
every mechanism our frozen pipeline admits, is at best indistinguishable from zero and at worst
negative (Table~\ref{tab:ablation}d).

\begin{figure}[!ht]\centering
\includegraphics[width=\textwidth]{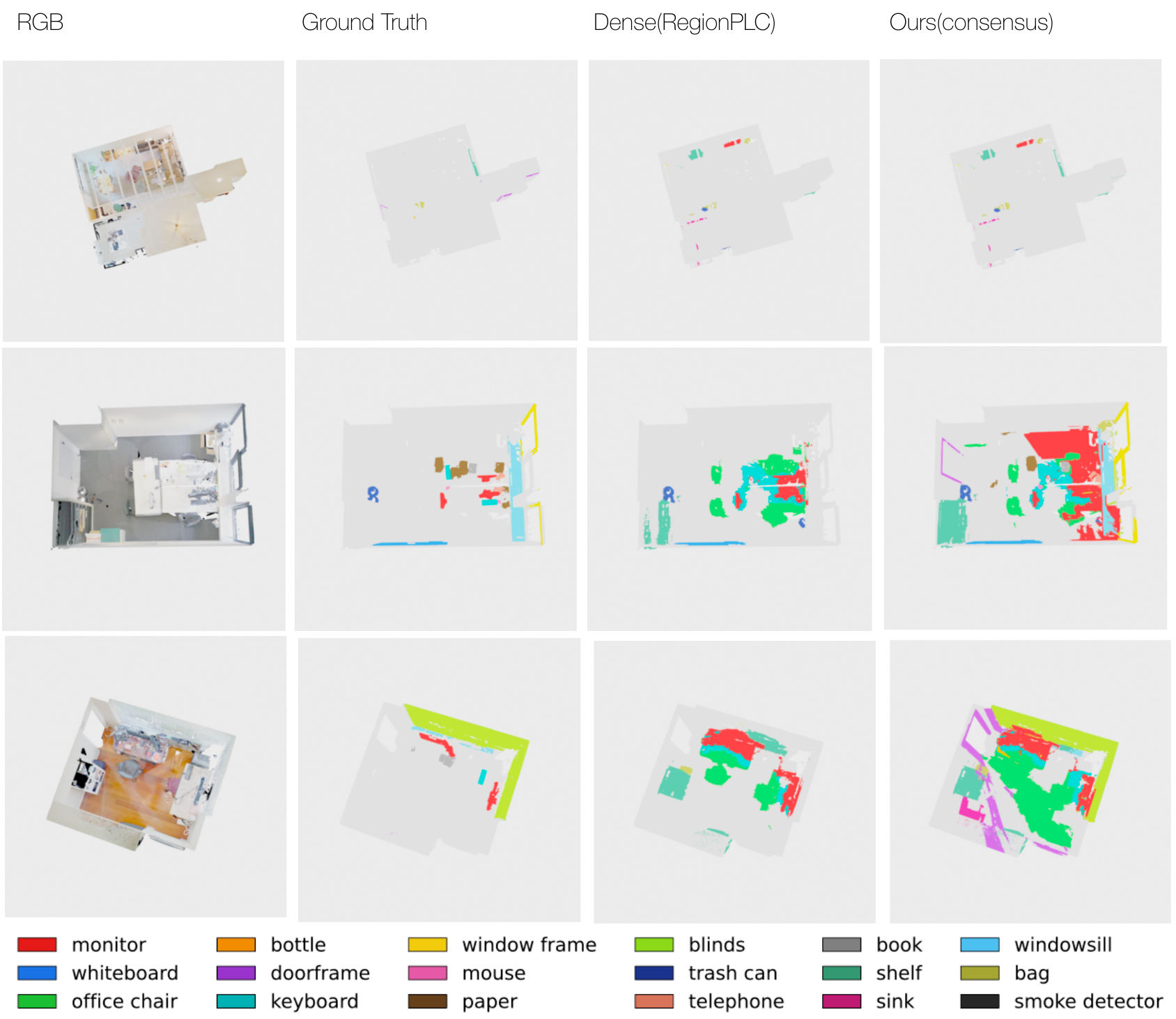}
\caption{\textbf{Qualitative results on ScanNet++} (novel classes colored, base and unlabeled points
grey; near-top-down view with the ceiling removed for visibility). Columns: input RGB, ground-truth
novel labels, the dense prior (RegionPLC), and Ours (cross-view consensus, $\tau{=}0.15$). Consensus
recovers small text-promptable novel objects the dense prior misses---the windowsill and window frame
(row~2) and the blinds strip (row~3)---at the cost of mild over-extension on a few classes, consistent
with the small base-mIoU cost we report. Per-scene novel mIoU (dense$\to$Ours):
$15.5{\to}23.0$, $19.5{\to}33.2$, $6.9{\to}17.2$.}
\label{fig:qual}
\end{figure}

\paragraph{Where the gain comes from.} The full per-class breakdown (supplementary Table~S1;
Fig.~\ref{fig:qual} shows the effect on three scenes) is sharp. The gain
comes almost entirely from small, text-promptable objects
that the dense prior misses outright: on ScanNet++, blinds ($+54.6$), windowsill ($+51.0$),
telephone, smoke detector and mouse each rise from a dense IoU of essentially zero to $34$--$57$, and
bottle and bag climb by $\sim\!19$. The same holds on ScanNet200, where pillow ($+29.1$), keyboard
($+24.2$), picture ($+19.6$), mirror ($+19.5$) and box ($+18.0$) lead. Conversely, the sparse branch
only ever \emph{hurts} a handful of large classes the dense prior already segments well---monitor
($-29.3$), office chair ($-12.4$) and whiteboard ($-7.2$) on ScanNet++; monitor, tv, couch and
armchair on ScanNet200---where a confused concept mask overwrites an already-correct dense label.
The net effect is strongly positive because the many near-zero classes have far more to gain than the
few strong classes have to lose, and it explains why consensus (rather than unconditional overwrite) is essential: it is precisely the strong-dense classes that a lower $\tau$ would damage. They are also the source of the $1.7$-point base cost in Table~\ref{tab:scannetpp}: the sparse branch's occasional overwrite of an already-correct dense label on these few classes is what the base mIoU reflects, not a systematic degradation across the base set.

\paragraph{Budget and runtime.} Because cross-view agreement is a \emph{fraction} of views, accuracy
degrades gracefully as posed frames are dropped: on ScanNet++, halving the frames ($16$/scene)
retains most of the gain (novel $31.9\!\to\!29.2$, HM $31.1\!\to\!29.7$) at half the lift cost, and
only past stride~$4$ does novel accuracy fall off; ScanNet200 is flatter still ($6$ frames/scene keep
novel at $24.2$ vs.\ $26.4$ at the full rate) because its dense video stream is redundant. Base
accuracy is flat or rises slightly as frames drop (fewer overwrites), so the frame rate is a clean
accuracy-for-cost dial. The pipeline is training-free and its per-scene cost is dominated by the two
frozen forward passes rather than the parameter-free reconciliation ($\sim\!20$\,s/scene on
ScanNet++, $\sim\!3$\,s on ScanNet200). The full frame-rate and runtime breakdowns are given in the
supplementary material.



\section{Discussion}\label{sec:discussion}

\paragraph{Why does injecting the support fail?} The failure is not that the support is
uninformative but that our frozen pipeline offers no \emph{well-calibrated} place to spend it.
The two loci it admits both route the support through RegionPLC features, and those features do not
live in the same calibrated space as the model's open-vocabulary text logits. The prototypical
blend (L2) adds a raw cosine similarity to a temperature-shaped softmax score; because the two
quantities are on different scales, even a small blend weight lets the flatter similarity term
dominate and collapses the dense novel score (Sec.~\ref{sec:support}), which is why the effect is
strictly negative rather than merely null. The metric gate (L1) avoids this by only \emph{filtering}
lifted votes, but cross-view consensus has already discarded the unreliable ones, so it has nothing
left to remove. The natural remedy---temperature-scaling or $\ell_2$/rank normalizing the prototype
logits into the text-logit space before blending---would very likely reduce the harm, but it
reintroduces a tuned, data-dependent parameter and thus the training-like calibration step the
setting is designed to avoid; we therefore report the negative result honestly and leave calibrated
support fusion, together with the exemplar-conditioned sparse locus (L3), as the open question. The
scoped claim stands: within a strictly frozen, training-free pipeline, the provided support has no
beneficial locus to enter once consensus is applied.

\paragraph{Label-list dependence and prompting.} Our only label-side input is the official novel
class string, used verbatim: we apply no synonyms, templated prompts, or prompt ensembles. This
keeps the comparison clean and the supervision minimal, but it means ambiguous or non-object names
(\eg ``object,'' ``ceiling,'' ``shower wall'') are poorly localized by the concept segmenter and are
carried entirely by the dense branch. Synonym or template ensembles are a known lever in 2D
open-vocabulary recognition and could raise novel accuracy on such names, at the risk of firing on
base surfaces; we leave this to future work. We use ``open-vocabulary'' in the standard 3D sense: 
arbitrary text labels accepted at inference without retraining, while noting, as the benchmark
does, that a label list is still provided; we do not claim generalization to free-form or unseen
concepts.

Two further points why a per-pixel CLIP/LSeg project-and-vote baseline is a \emph{weaker} instance
of our RegionPLC dense prior rather than an independent axis, and the method's dependence on posed
imagery and camera calibration are discussed in the supplementary material.

\section{Conclusion}
We asked how far the dense--sparse reconciliation at the heart of GFS-PCS
can be pushed with no training, no 3D labels, and not even the few-shot
support. Our answer is a parameter-free pipeline: a frozen 3D
vision-language dense prior (RegionPLC) reconciled with a frozen
promptable concept segmenter (SAM3) by a single cross-view consistency
threshold that relabels a point novel only when enough of the views that
see it agree, suppressing the single-view false positives that make naive
overwrite trade base accuracy for novel. This raises novel mIoU by $+2.6$
at a $0.5$ base cost on ScanNet200 (a third of the gap to trained GFS-VL)
and nearly doubles it on ScanNet\texttt{++} ($+15.7$, $16.2 \to 31.9$;
HM $21.5 \to 31.1$), bringing the harmonic mean to within $0.7$ of
trained GFS-VL despite a substantial remaining base-class gap, the
standing cost of forgoing base 3D labels and training. Our ablations show
consistency is the whole story: visibility and dense-confidence gates add
nothing, and injecting the provided support through a fusion gate or a
prototypical classifier is null or negative. Conditioned on consensus the
support carries no beneficial signal, which licenses the support-free,
open-vocabulary setting and establishes a training-free floor for the
benchmark.

\paragraph{Limitations and future work.}
Forgoing base 3D labels leaves a standing base-class gap to the trained
state of the art. The sparse branch only helps classes a text-promptable
segmenter can name and localize: object-like novel classes are lifted,
while stuff/abstract classes (e.g.\ ceiling, doorframe) are carried
entirely by the dense prior, and $\tau$ must be calibrated once per
benchmark. Both support-injection loci we test route the support through
RegionPLC features, so a shared-feature explanation cannot be fully
excluded; the exemplar-conditioned locus (L3) that would settle it, and
swapping in a stronger frozen dense prior to raise the floor, are left to future work.

\bibliographystyle{splncs04}
\bibliography{main}

\end{document}